# Decoupling Learning Rules from Representations

**Philip S. Thomas**
Carnegie Mellon University

**Christoph Dann**
Carnegie Mellon University

**Emma Brunskill**
Stanford University

## Abstract

In the artificial intelligence field, *learning* often corresponds to changing the parameters of a parameterized function. A *learning rule* is an algorithm or mathematical expression that specifies precisely how the parameters should be changed. When creating an artificial intelligence system, we must make two decisions: what representation should be used (i.e., what parameterized function should be used) and what learning rule should be used to search through the resulting set of representable functions. Using most learning rules, these two decisions are coupled in a subtle (and often unintentional) way. That is, using the same learning rule with two different representations that can represent the same sets of functions can result in two different outcomes. After arguing that this coupling is undesirable, particularly when using artificial neural networks, we present a method for partially decoupling these two decisions for a broad class of learning rules that span unsupervised learning, reinforcement learning, and supervised learning.

## 1 Introduction

Consider two challenges at the foundation of *artificial intelligence* (AI) research and practice: representation selection and learning rule selection. *Representation selection* is the decision of how knowledge should be represented within an AI system. *Learning rule selection* is the decision of which algorithm should be used to modify the system's stored knowledge. These two challenges are central to most AI systems, including unsupervised, supervised, and reinforcement learning systems.

Representation and learning rule selection are intertwined—a learning rule can work well with some representations and work poorly with, or be incompatible with, others. Although this intertwining is unavoidable and is likely desirable, it can have unintended and undesirable effects. To see this, we split representation selection into two components: deciding *what* the system should be able to represent (e.g., normal distributions) and *how* the system will represent it (e.g., by storing the mean and standard deviation or variance). The intertwining of the learning rule with the decision of *what* the system should be able to represent is often considered by the designer of an AI system. However, the intertwining of the learning rule with the decision of *how* the system will represent knowledge is not necessarily desirable, is often overlooked, and can have significant ramifications.

Consider an example, which we adapt from an example presented by Amari [1], where part of an AI system approximates an unknown distribution that generated some observed data, $X_1, X_2, \ldots, X_n$, where each $X_i \in \mathbb{R}$. We might decide that the system will estimate the distribution using a normal distribution (*what* the system can represent) and that the normal distribution will be represented by storing its mean, $\mu$, and standard deviation, $\sigma$ (*how* the system represents it). We might then decide to use gradient descent with step size $\alpha \coloneqq .001/n$ to maximize the log-likelihood of the model, i.e.,

$$\mu_{i+1} = \mu_i + \frac{\alpha}{\sigma_i^2} \sum_{j=1}^n (X_j - \mu_i) \quad \text{and} \quad \sigma_{i+1} = \sigma_i - \frac{\alpha n}{\sigma_i} + \frac{\alpha}{\sigma_i^3} \sum_{j=1}^n (X_j - \mu_i)^2.$$

Clearly in this setting the decision to model the distribution that generated the data using a normal distribution will impact the behavior of the resulting system, and this decision would likely be carefully considered. Less obviously, our decision to parameterimize normal distributions using $\mu$ and $\sigma$ will also impact the behavior of the system. If we chose to parameterize normal distributions using the variance, $\sigma^2$ rather than the standard deviation, $\sigma$, our same learning rule would produce a different sequence of normal distributions. In general, if we represent normal distributions with two parameters, $\mu$ and $\sigma^k$, the resulting stochastic gradient descent updates are:

$$\mu_{i+1} = \mu_i + \frac{\alpha}{(\sigma_i^k)^{\frac{2}{k}}} \sum_{j=1}^{n}(X_j - \mu_i) \quad \text{and} \quad \sigma_{i+1}^k = \sigma_i^k - \frac{\alpha n}{k \sigma_i^k} + \frac{\alpha}{k}(\sigma_i^k)^{-\frac{k+2}{k}} \sum_{j=1}^{n}(X_j - \mu_i)^2.$$

Figure 1 shows the results of applying this algorithm to a fixed data set using various $k$. Notice that different choices of how to represent normal distributions result in wildly different outcomes. A poor choice can result in a sequence of normal distributions that takes a circuitous path to the maximum log-likelihood distribution and produces poorly scaled updates. As a result, a poor choice of *how* to represent normal distributions can result in the likelihood of the model increasing slowly (notice that using $\sigma^4$, the model failed to approach the maximum log-likelihood model within 200,000 iterations).

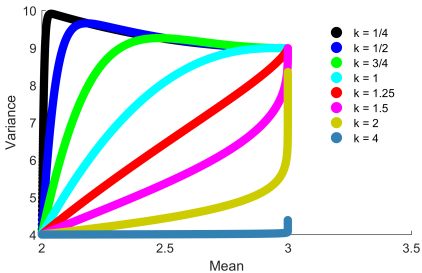

Figure 1: We generated a data set containing 100,000 samples from $\mathcal{N}(3, 9)$—the normal distribution with mean 3 and variance 9. Each curve shows the sequence of normal distributions produced by using gradient descent to maximize the log-likelihood of a parameterized normal distribution for 200,000 iterations and starting from $\mathcal{N}(2, 4)$. The normal distributions are parameterized by $\mu$, their mean, and $\sigma^k$, a power of their standard deviation. Each curve therefore corresponds to using a different (but equally representative) representation, but the same learning rule.

In this case, and many others, the underlying cause of the intertwining of the learning rule with the choice of how to represent knowledge stems from an implicit assumption hidden within the learning rule (stochastic gradient descent in this case): distances between different parameter vectors, $\theta$, should be measured using Euclidean distance. In our example, $\theta = [\mu, \sigma^k]^\intercal$, and the distance between $\theta = [\mu, \sigma^k]^\intercal$ and $\theta' = [\mu', \sigma'^k]^\intercal$—the distance between $\mathcal{N}(\mu, \sigma^2)$ and $\mathcal{N}(\mu', \sigma'^2)$—is:

$$\text{dist}(\theta, \theta') \coloneqq \sqrt{(\theta' - \theta)^\intercal (\theta' - \theta)} = \sqrt{(\mu' - \mu)^2 + (\sigma'^k - \sigma^k)^2}.$$

Notice that this definition of distance is dependent on $k$, or, more generally, on *how* the parameter vector encodes knowledge. In this case, large $k$ (e.g., $k = 4$) result in small changes to $\sigma$ incurring large amounts of distance relative to similar changes to $\mu$ (given that $\sigma > 1$). As a result, using large $k$ results in a sequence of normal distributions that focuses on changing the mean first, since changes to $\mu$ incur less distance than changes to $\sigma$. Similarly, small $k$ results in a sequence of normal distributions that over-emphasizes adjusting the variance of the normal distribution.

Sometimes it is easy to control *how* knowledge is represented, like when deciding how to represent normal distributions in our example, and it may also be easy to select a representation such that Euclidean distance in the parameters is reasonable (e.g., using $\sigma$ or $\sigma^2$, but not $\sigma^{20}$). However, in other cases it may not be clear how to define $\theta$ so that it can represent what we want while simultaneously ensuring that Euclidean distance between parameter vectors is a reasonable notion of distance.

This is particularly true when using deep *artificial neural networks* (ANNs), where the parameter vector, $\theta$, corresponds to the weights of the ANN. If a weight near the output layer has a bigger impact on the function represented by the ANN than a weight in an early layer, then using Euclidean distance between weight vectors $\theta$ and $\theta'$ to measure the distance between the functions represented by an ANN with weights $\theta$ and $\theta'$ will place undue emphasis on weights near the output layer. For ANNs, this symptom of the underlying problem is called the problem of vanishing gradients [2].

In this paper we study the intertwining of the decision of which learning rule to use and *how* knowledge should be represented, extending works by Amari [1], Kakade [3], and Bagnell and Schneider [4]. For a broad class of gradient-like learning rules, which subsumes gradient, stochastic gradient, and accelerated gradient methods as well as temporal-difference methods [5], we present a method for modifying the learning rule to use an explicit definition of how distances should be measured during learning rather than requiring the use of Euclidean distance between parameter vectors. Importantly, this method is *not* a learning rule itself, but rather a method that other researchers can use to enhance the learning rules that they design.

More specifically, we begin by defining different forms of *covariance*, which capture different levels with which a learning rule can be independent of the choice of how knowledge will be represented. We then propose a method for converting any learning rule from within a broad class of learning rules into a first-order covariant learning rule—the weakest form of decoupling—and show how this method can be approximated without increasing the computational complexity of the learning rule. Our method is closely related to natural gradient methods [6]—it extends them to a more general class of learning rules. We also discuss why a second-order covariant learning rule would be desirable before proving that no useful second-order covariant learning rules exist. We conclude with examples of how some existing learning rules can be converted into first-order covariant learning rules and discuss some of the subtleties and surprising properties of our method.

## 2 Notation and Definitions

Let a *parameter vector*, $\theta$, be a vector that captures the current knowledge of an AI system. Let $\Theta \subseteq \mathbb{R}^n$ be the set of all possible values for $\theta$. We restrict our discussion to AI systems that use a parameter vector to parameterize a function. Intuitively, a *parameterized function* is a function that, for each possible parameter vector, $\theta \in \Theta$, produces a mapping from elements of some set, $\mathcal{X}$, to elements of $\mathbb{R}^k$. More formally, $f : \mathcal{X} \times \Theta \to \mathbb{R}^k$ so that $f(x, \theta)$ denotes the parameterized function evaluated at $x \in \mathcal{X}$ using the parameter vector $\theta \in \Theta$. For example, in a regression system $f(x, \theta)$ might be an estimate of $f^\star(x)$ for some target function $f^\star : \mathcal{X} \to \mathbb{R}^k$ and some $x \in \mathcal{X}$. Similarly, in a reinforcement learning system $f(x, \theta)$ might be a scalar estimate of the value, $q^\pi(x)$, of a state-action pair, $x$, under some policy, $\pi$. Let $\mathcal{P}$ be the set of all parameterized functions.

Importantly, we further restrict our discussion to systems where the parameterized function is in $C^1$: we only consider parameterized functions, $f$, such that for all $x \in \mathcal{X}$ and $\theta \in \Theta$, $\frac{\partial f}{\partial \theta}(x, \theta)$ exists. Although this assumption rules out some AI systems, like ones that use STRIPS [7] or ID3 [8], it applies to many systems like those that use linear estimators or ANNs.

*Learning* is the iterative search for parameter vectors that cause $f(\cdot, \theta)$ to achieve some desirable property, starting from some set of initial parameter vectors, $\theta_0 \coloneqq (\theta_0^1, \theta_0^2, \ldots, \theta_0^\iota) \in \Theta^\iota$, where $\iota \in \mathbb{N}_{\geq 0}$ denotes the number of initial parameter vectors required by the learning rule. Although $\iota = 1$ for most learning rules, like gradient descent, some learning rules (like the accelerated gradient methods discussed later) require multiple initial parameter vectors.

Let $(\Omega, \Sigma, p)$ be a probability space that captures all sources of randomness that occur during the learning process. Intuitively, each outcome, $\omega \in \Omega$, is a seed for the random number generator used by both the AI system and the environment with which it interacts. A *learning rule*, $l$, is a sequence of functions, $l \coloneqq (l_i)_{i=1}^\infty$ where each $l_i : \mathcal{P} \times \Theta^\iota \times \Omega \to \Theta$ such that $l_i(f, \theta_0, \omega)$ denotes the parameter vector at the $i^\text{th}$ iteration when learning rule $l$ is used with the parameterized function $f$, the initial parameter vectors are $\theta_0 \in \Theta^\iota$, and all randomness is captured by the outcome $\omega \in \Omega$. To simplify notation for the base cases in recursive expressions, when $i \leq 0$ let $l_i(f, \theta_0, \omega) \coloneqq \theta_0^{1-i}$. For example, consider the learning rule, $l$, for stochastic gradient descent when minimizing the expected squared error between $f(\cdot, \theta)$ and a target function, $f^\star : \mathcal{X} \to \mathbb{R}$:

$$l_{i+1}(f, \theta_0, \omega) = l_i(f, \theta_0, \omega) - \alpha_i \Big(f^\star(X_i(\omega)) - f(X_i(\omega), l_i(f, \theta_0, \omega))\Big) \frac{\partial f}{\partial l_i(f, \theta_0, \omega)}(X_i(\omega), l_i(f, \theta_0, \omega)),$$

where $X_i : \Omega \to \mathcal{X}$ is the input to $f^\star$ used during the $i^\text{th}$ iteration, which is a random variable, and $(\alpha_i)_{i=1}^\infty$ is a sequence of positive real-valued step sizes. To obtain a more familiar notation, let $\theta_i \coloneqq l_i(f, \theta_0, \omega)$, which gives the definition: $\theta_{i+1} = \theta_i - \alpha_i \big(f^\star(X_i(\omega)) - f(X_i(\omega), \theta_i)\big) \frac{\partial f}{\partial \theta_i}(X_i(\omega), \theta_i)$.

Intuitively, we say that a parameterized function, $g$, is congruent to a parameterized function, $f$, if $g$ can represent everything that $f$ can, and there is a smooth mapping from parameters for $f$ to congruent parameters for $g$. More formally:

**Definition 1** (Congruent Representations). *Let $g : \mathcal{X} \times \Psi \to \mathbb{R}^k$ and $f : \mathcal{X} \times \Theta \to \mathbb{R}^k$ be two parameterized functions, where $\Psi \subseteq \mathbb{R}^m$ and $\Theta \subseteq \mathbb{R}^n$. We say that $g$ is congruent to $f$ if there* *exists a function* $\psi : \Theta \to \Psi$, *called a submersion, such that* $f(x, \theta) = g(x, \psi(\theta))$ *for all* $x \in \mathcal{X}$ *and* $\theta \in \Theta$, $\frac{\partial \psi}{\partial \theta}(\theta)$ *is full rank for all* $\theta \in \Theta$, *and* $m \leq n$.

Hereafter we reserve the symbols $f$, $g$, $\psi$, $\Theta$, and $\Psi$ to represent a parameterized function, $g : \mathcal{X} \times \Psi \to \mathbb{R}^k$, that is congruent to a parameterized function, $f : \mathcal{X} \times \Theta \to \mathbb{R}^k$, with submersion $\psi$, and we reserve the symbols $n$ and $m$ for the dimensions of $\Theta$ and $\Psi$, respectively. Also, notice that the congruency of functions is *not* symmetric. For example, if $g(x, \theta) = \theta$ and $f(x, \theta) = e^\theta$, where $\theta \in \mathbb{R}$, then $g$ is congruent to $f$, but $f$ is not congruent to $g$.

We can now state some earlier concepts more formally. The question of *what* an AI system should be able to represent is the question of what functions should be representable by the parameterized function that is used. The question of *how* an AI system should represent knowledge is the question of which parameterized function to use from a set of mutually congruent parameterized functions.

Intuitively, our goal is to ensure that a learning rule will produce the same sequence of functions regardless of which parameterized function is selected from a set of mutually congruent parameterized functions. Learning rules with this property are sometimes called *invariant to reparameterization*, since they do not change their behavior if the parameterized function that they use is parameterized in a different way. They are also sometimes called *covariant*, perhaps in reference to *covariant transformations* in physics. This usage of the word covariant has been attributed [9] to Knuth [10]. To maintain consistency with prior work in machine learning [3, 4, 6, 9, 11–14], and for brevity, we adopt the latter terminology. Formally, we define a covariant learning rule as follows (where $\psi(\theta_0) \coloneqq (\psi(\theta_0^1), \psi(\theta_0^2), \ldots, \psi(\theta_0^\iota)))$:

**Definition 2** (Covariant Learning Rule). *A learning rule, $l$, is a covariant learning rule if for all parameterized functions, $f$, all parameterized functions $g$ that are congruent to $f$, all $i \in \mathbb{N}_{>0}$, all $\omega \in \Omega$, all $\theta_0 \in \Theta^\iota$, and all $x \in \mathcal{X}$, $f(x, l_i(f, \theta_0, \omega)) = g(x, l_i(g, \psi(\theta_0), \omega))$.*

Although covariant learning rules exist, they tend to be computationally impractical and are therefore rare. Instead of requiring a learning rule to produce the exact same sequence of functions, $(f(\cdot, l_i(f, \theta_0, \omega)))_{i=1}^\infty$ regardless of the parameterized function that is used, we can require a learning rule to produce a sequence of functions that, at each iteration, changes the parameter vector in a direction that is locally independent of the parameterization. More specifically, we define a $j$-order covariant learning rule to be a learning rule that ensures that a $j$-order Taylor approximation of $f(\cdot, l_i(f, \theta_0, \omega))$, centered around $l_{i-1}(f, \theta_0, \omega)$, is independent of $f$:

**Definition 3** ($j$-Order Covariant Learning Rule With Respect to a Set, $\mathcal{G}$). *A learning rule, $l$, is a $j$-order covariant learning rule with respect to a set $\mathcal{G} \subseteq \mathcal{P}$ if for all parameterized functions, $f \in \mathcal{P}$, all $g \in \mathcal{G}$ that are congruent to $f$, all $i \in \mathbb{N}_{>0}$, all $\omega \in \Omega$, all $\theta_0 \in \Theta^\iota$, and all $x \in \mathcal{X}$,*

$$\tau_j\left(f(x,\cdot), l_{i-1}(f, \theta_0, \omega), l_i(f, \theta_0, \omega)\right) = \tau_j\left(g(x,\cdot), \psi(l_{i-1}(f, \theta_0, \omega)), l_i(g, \psi(\theta_0), \omega)\right),$$

*where $\tau_j(h, y, y')$ denotes the $j$-order Taylor approximation of $h(y')$ centered around $h(y)$.*

Notice that a covariant learning rule is *not* necessarily $j$-order covariant for any $j$, nor does $j$-order covariance imply $(j-k)$-order covariance for any $k > 0$. Also, although $j$-order covariance captures different levels of covariance, it is limited to learning rules of the form $\theta_i = \theta_{i-1} + \alpha_i \Delta_i$, where $\Delta_i \in \mathbb{R}^n$ is the update direction at step $i$. That is: it is restricted to updates that take a step from the previous parameters, $\theta_{i-1}$. Some learning rules, like accelerated gradient methods [15], take steps from some other point: $\theta_i = \beta_i + \alpha_i \Delta_i$, where $\beta_i$ might be some combination of previous parameter vectors, like $\beta_i = \gamma \theta_{i-1} + (1-\gamma)\theta_{i-2}$, where $\gamma \in [0, 1]$. To provide an achievable form of covariance for these updates, we require them to be equivalent under a $j$-order Taylor approximation centered around $\beta_i$. This leads to the more general definition of $j$-order covariance:

**Definition 4** ($j$-Order Covariant Learning Rule with Respect to a Sequence and Set). *Let $\beta_1, \beta_2, \ldots$ be a sequence where each $\beta_i \in \mathbb{R}^n$. A learning rule, $l$, is a $j$-order covariant learning rule with respect to the sequence $(\beta_i)_{i=1}^\infty$ and a set $\mathcal{G} \subseteq \mathcal{P}$, if for all parameterized functions, $f \in \mathcal{P}$, all $g \in \mathcal{G}$ that are congruent to $f$, all $i \in \mathbb{N}_{>0}$, all $\omega \in \Omega$, all $\theta_0 \in \Theta^\iota$, and all $x \in \mathcal{X}$,*

$$\tau_j\left(f(x,\cdot), \beta_i, l_i(f, \theta_0, \omega)\right) = \tau_j\left(g(x,\cdot), \psi(\beta_i), l_i(g, \psi(\theta_0), \omega)\right).$$

This is *not* consistent with previous definitions of a covariant learning rule [4, 12–14]—what these previous authors called "covariant" learning rules we call "first-order covariant". We adopt our alternate definition because later we discuss higher orders of covariance.

Furthermore, previous works refer to covariant updates without specifying the set, $\mathcal{G}$, that the covariance is with respect to—we make this set explicit. As an example, Thomas et. al [14] provide a proof of first-order covariance for a specific learning rule that requires the implicit assumption that $\mathcal{G}$ only includes $g$ with positive definite *energetic information matrices*. Similarly, Amari [1, page 16] restricts $f$ and $g$ to parameterized discrete probability distributions for which the *Fisher information matrix* is positive definite (it is in general only guaranteed to be positive semi-definite). We make these restrictions on $g$ explicit in the definition of $j$-order covariant learning rules. Notice that the the strength of the $j$-order covariance property scales not just with $j$, but also with the size of $\mathcal{G}$.

Next we will define a *covariant signed measure*, which we will use when defining the class of learning rules that we can transform into first-order covariant learning rules. Let $\mathcal{F}$ be a $\sigma$-algebra on $\mathcal{X}$. For all $i \in \mathbb{N}_{>0}$, let $\mu_i : \mathcal{P} \times \Theta^\iota \times \Omega \times \mathcal{F} \to \mathbb{R}$ such that $(\mathcal{X}, \mathcal{F}, \mu_i(f, \theta_0, \omega, \cdot))$ is a signed measure space for all $f \in \mathcal{P}$, all $\theta_0 \in \Theta^\iota$, and all $\omega \in \Omega$.

**Definition 5** (Covariant Signed Measures). *We say that $\mu = (\mu_i)_{i=1}^\infty$ is a covariant set of signed measures if for all $i \in \mathbb{N}_{>0}$, all $f \in \mathcal{P}$, all $g$ congruent to $f$, all $\theta_0 \in \Theta^\iota$, all $\omega \in \Omega$, and all $E \in \mathcal{F}$, $\mu_i(f, \theta_0, \omega, E) = \mu_i(g, \psi(\theta_0), \omega, E)$.*

Lastly, we define a *covariant joint probability measure*, which we will use when defining a notion of distances between functions:

**Definition 6** (Covariant Joint Probability Measures). *We say that a set, $p = (p_i)_{i=1}^\infty$ is a covariant set of joint probability measures if for all $i \in \mathbb{N}_{>0}$, all $x \in \mathcal{X}$, all $f \in \mathcal{P}$, all $g$ congruent to $f$, all $\theta_0 \in \Theta^\iota$, all and $\omega \in \Omega$, $(\mathcal{X}^2, \mathcal{F}^2, p_i(f, \theta_0, \omega, x, \cdot, \cdot))$ is a probability space, and for all $(E, E') \in \mathcal{F}^2$, $p_i(f, \theta_0, \omega, x, E, E') = p_i(g, \psi(\theta_0), \omega, x, E, E')$.*

## 3 How to Make a Learning Rule First-Order Covariant

In Theorem 1, presented later in this section, we show how a broad class of learning rules can be transformed into first-order covariant learning rules. Intuitively, the $\partial f(x, \theta)/\partial \theta$ terms within a learning rule make the implicit assumption that distances between parameterized functions, $f(\cdot, \theta)$ and $f(\cdot, \theta')$, should be measured using Euclidean distance between their parameter vectors. This implicit assumption has been described previously [6]. In Appendix B we review exactly where this implicit assumption is made. We propose replacing the $\partial f(x, \theta)/\partial \theta$ terms within a learning rule with terms that respect a user-defined distance. Furthermore, we allow different distance measures over functions to be used when replacing each $\partial f(x, \theta)/\partial \theta$ term in a learning rule.

Theorem 1 is closely related to *natural gradient* methods [6]. Amari [6] argued that, if **1)** the goal is to minimize a loss function, $L : \mathbb{R}^n \to \mathbb{R}$, **2)** the arguments (inputs) of $L$ lie on a Riemannian manifold characterized by the positive-definite metric tensor $G : \Theta \to \mathbb{R}^{n \times n}$, and **3)** one plans to use a gradient descent algorithm of the form $\theta_{i+1} \leftarrow \theta_i - \alpha_i \nabla L(\theta_i)$, then one should instead use the *natural gradient* update $\theta_{i+1} \leftarrow \theta_i - \alpha G(\theta_i)^{-1} \nabla L(\theta_i)$. Thomas [16] generalized the natural gradient to allow the parameters of $L$ to lie on a semi-Riemannian manifold and established sufficient conditions for the convergence of natural gradient descent (which differ from those of ordinary gradient descent), but did not establish covariance properties.

Intuitively, one can view our approach as replacing gradient terms, $\partial f(x, \theta)/\partial \theta$, within a learning rule, with (generalized) natural gradients, $G(\theta)^+ \partial f(x, \theta)/\partial \theta$, where $G(\theta)$ is automatically derived from a user-provided notion of the distance between parameterized functions. Unlike Amari's original natural gradient method, the gradient terms that we replace with natural gradient terms are *not* the gradients of a loss function, but rather terms within a learning rule. In Appendix B we show how the $G(\theta)$ that we use can be derived from the provided distance measure and review Amari's argument for why using $G(\theta)^+ \partial f(\cdot, \theta)/\partial \theta$ terms in place of $\partial f(x, \theta)/\partial \theta$ terms corresponds to using the provided distance function rather than Euclidean distance. Due to the strong influence that Amari's work with natural gradients had on our approach, we refer to the learning rule, $\tilde{l}$, produced by applying our method to a learning rule $l$, as a *naturalized* version of $l$. Also, the naturalized version of gradient descent will be Amari's natural gradient algorithm.

The implicit distance (which may not satisfy the requirements of a measure, and so technically it is a dissimilarity function, not a distance measure) used by the naturalized learning rule takes the form:

$$\text{dist}(\theta, \theta + \Delta) = \sqrt{\frac{1}{2} \int_{\mathcal{X}^2} (f(x,\theta) - f(x,\theta+\Delta))(f(y,\theta) - f(y,\theta+\Delta)p(dx,dy)}, \quad (1)$$

where the designer of the learning rule that is being converted into a first-order covariant rule is free to select the joint probability measure, $p$, to capture the desired notion of distance. We will allow for different probability measures, $p$, for each $\partial f(x,\theta)/\partial\theta$ term in the learning rule, and we will write $p_i(f, \theta_0, \omega, x, \cdot, \cdot)$ to denote the probability measure used during the $i^{\text{th}}$ update for terms associated with $x$. Usually we will define $p$ so that $x = y$, and $p$ will correspond to a probability mass function (or probability density function), in which case we can write the notion of distance as:

$$\text{dist}(\theta, \theta + \Delta) = \sqrt{\frac{1}{2} \sum_{x \in \mathcal{X}} p(x)(f(x,\theta) - f(x, \theta + \Delta))^2}.$$

This makes it clear that our notion of distance between $f(\cdot, \theta)$ and $f(\cdot, \theta + \Delta)$ is the expected squared difference between their values given that the inputs, $x$, are sampled from some distribution, $p(\cdot)$, which is chosen by the designer of the algorithm.

The learning rules that our method can transform into first-order covariant learning rules satisfy:

**Assumption 1.** *The learning rule, $l$, can be written as:*

$$l_i(f, \theta_0, \omega) = l'_i(f, \theta_0, \omega) + \int_{\mathcal{X}} \frac{\partial f}{\partial \beta_i}(\cdot, \beta_i) \, d\mu_i(f, \theta_0, \omega, \cdot),$$

*where $l'$ is a first-order covariant learning rule with respect to some sequence $(\beta_i)_{i=1}^\infty$ and $\mu$ is a covariant set of signed measures.*

Notice that $l'_i(f, \theta_0, \omega) \coloneqq \beta_i$ is a first order covariant learning rule with respect to $(\beta_i)_{i=1}^\infty$, and that this definition of $l'$ will be the most common. We are now ready to present our main theorem, which takes a learning rule, $l$, and transforms it into a new learning rule, $\tilde l$, that is first-order covariant.

**Theorem 1.** *Given a learning rule, $l$, that satisfies Assumption 1, a covariant set of joint probability measures, $p$, and a covariant set of measures, $\mu$, the learning rule $\tilde l$ defined by:*

$$\tilde l_i(f, \theta_0, \omega) = l'_i(f, \theta_0, \omega) + \int_{\mathcal{X}} G_i^{\cdot, \theta_0, \omega}(f, \beta_i)^+ \frac{\partial f}{\partial \beta_i}(\cdot, \beta_i) \, d\mu_i(f, \theta_0, \omega, \cdot),$$

$$G_i^{z, \theta_0, \omega}(f, \beta_i) \coloneqq \int_{\mathcal{X}^2} \frac{\partial f}{\partial \beta_i}(x, \beta_i) \left(\frac{\partial f}{\partial \beta_i}(y, \beta_i)\right)^{\intercal} dp_i(f, \theta_0, \omega, z, dx, dy), \quad (2)$$

*is a first-order covariant learning rule with respect to $(\beta_i)_{i=1}^\infty$ and $\mathcal{G}$, where $\mathcal{G}$ is the set of parameterized functions, $g \subseteq \mathcal{P}$ such that $G_i^{z, \theta_0, \omega}(g, \beta_i)$ is full rank for all $z \in \text{supp}(\mu_i(f, \theta_0, \omega, \cdot))$.*
**Proof.** *See Appendix A.* □

## 4 Direct Estimation of the Update

In cases where $G_i^{x, \theta_0, \omega}(f, \beta_i)$ is not sparse, $\tilde l$ can have high computational complexity—$O(n^3)$ for naïve implementations. In this section we show that in some cases $\tilde l$ can be estimated directly without even requiring the estimation of the $n \times n$ matrix $G_i^{x, \theta_0, \omega}(f, \beta_i)$. Whereas Theorem 1 was inspired by Amari's work with natural gradients, the linear time approximation presented here generalizes Kakade's work [3] and Bhatnagar et al.'s work [17] showing that the natural policy gradient in reinforcement learning can be estimated in linear time by using *compatible function approximation* [18]. Let $\hat 1 : \mathcal{X} \times \mathbb{R}^n$ be a parameterized function defined by $\hat 1(x, w) \coloneqq w^{\intercal} \frac{\partial f}{\partial \beta_i}(x, \beta_i)$, where $k = 1$ so that $f(x, \theta) \in \mathbb{R}$.

**Theorem 2.** *If $k=1$, $w^\star \in \arg\min_{w \in \mathbb{R}^n} \int_{\mathcal{X}} \left(1 - \hat 1(\cdot, w)\right)^2 d\mu_i(f, \theta_0, \omega, \cdot)$, and $G_i^{x, \theta_0, \omega}(f, \beta_i) = \int_{\mathcal{X}} \frac{\partial f}{\partial \beta_i}(\cdot, \beta_i) \frac{\partial f}{\partial \beta_i}(\cdot, \beta_i)^{\intercal} d\mu_i(f, \theta_0, \omega, \cdot)$ or $G_i^{x, \theta_0, \omega}(f, \beta_i) = \frac{\partial f}{\partial \beta_i}(x, \beta_i) \frac{\partial f}{\partial \beta_i}(x, \beta_i)^{\intercal}$, and $G_i^{x, \theta_0, \omega}(f, \beta_i)$ is full rank for all $\beta_i$ and the single $f$ that is being used, then the learning rule, $\tilde l$, in Theorem 1 can be written as $\tilde l_i(f, \theta_0, \omega) = l'_i(f, \theta_0, \omega) + w^\star$.*
**Proof.** *See Appendix C.* □

Intuitively, Theorem 2 says that if $w^\star \in \mathbb{R}$ are parameters for $\hat{1}$ that minimize the average difference between $\hat{1}(x, w)$ and 1, weighted by the signed measure $\mu_i$, then $\tilde{l}$ takes a step in the direction $w^\star$ from $\beta_i$. If $w^\star$ can be efficiently estimated, then it may be more computationally efficient to estimate $w^\star$ than it is to estimate $G_i^{x,\theta_0,\omega}$ and compute the product of its pseudoinverse with $\partial f(x, \beta_i)/\partial \beta_i$.

For example, in the context of natural policy gradient methods for reinforcement learning, Bhatnagar et. al [17] suggest using a two-timescale approach to simultaneously estimate $w^\star$ using stochastic gradient descent on $\int_{\mathcal{X}} \left(1 - \hat{1}(\cdot, w)\right)^2 d\mu_i(f, \theta_0, \omega, \cdot)$ and update the parameters $\theta$ in the direction $w^\star$. This approach results in computational complexity $O(n)$ per time step, since the stochastic gradient update for estimating $w^\star$ takes $O(n)$ time. However, notice that methods of this form only *approximate* a a first-order covariant learning rule, because an estimate of $w^\star$ is used instead of $w^\star$.

## 5 The Non-Existence of Second-Order Covariant Learning Rules

While investigating the proper use of Amari's natural gradient methods for policy search in reinforcement learning, Bagnell and Schneider [4], noticed that a natural gradient method (using the Fisher information matrix for $G(\theta)$, which results in a first-order covariant update with respect to the set of parameterized probability distributions that have positive definite Fisher information matrices) did not act in a covariant way in practice. They concluded that this was due to their use of large step sizes, since first-order covariant learning rules will only behave in a covariant manner for step sizes that are sufficiently small for the first-order Taylor approximation to be accurate. This raises the question: can one develop a second-order covariant learning rule? If such a learning rule existed, then it might behave in a covariant manner when using larger step sizes.

Although we set out to construct a second-order covariant learning rule, we were unable to find any for non-degenerate $\mathcal{G}$, other than the trivial learning rule $l_i(f, \theta_0 \omega) := \beta_i$. In Theorem 3 we give a one dimensional (i.e., $n = m = k = 1$) example of a reasonable class of $\mathcal{G}$ for which no second-order covariant learning rules exist. We conjecture that no second order learning rules exist for a far broader class of similar $\mathcal{G}$. We say that two functions, $\rho$ and $\varrho$, both with domain $\mathcal{X}$, are *collinear* if there exists a constant $\gamma \in \mathbb{R}$ such that for all $x \in \mathcal{X}$, $\rho(x) = \gamma \varrho(x)$.

**Theorem 3** (Nonexistence of Nontrivial Second-Order Covariant Learning Rules). *Every learning rule, l, that is second-order covariant with respect to any sequence, $(\beta_i)_{i=1}^\infty$, and a set, $\mathcal{G}$, must use the trivial update, $l_i(f, \theta_0, \omega) := \beta_i$ for all parameterized functions, $f$, where* **1)** *$n = k = 1$,* **2)** *both $g(x, \theta) := f(x, \ln(\theta))$ and $h(x, \theta) := f(x, \ln(\theta)/2)$ are in $\mathcal{G}$ and are congruent to $f$ and* **3)** *both $\frac{\partial g}{\partial \theta}(\cdot, \beta_i)$ and $\frac{\partial^2 g}{\partial \theta^2}(\cdot, \beta_i)$ are not collinear and $\frac{\partial h}{\partial \theta}(\cdot, \beta_i)$ and $\frac{\partial^2 h}{\partial \theta^2}(\cdot, \beta_i)$ are not collinear.*
**Proof.** *See Appendix D.* □

## 6 Discussion and Conclusion

First, notice that Theorem 1 generalizes several existing results. If $\mathcal{P}$ contains parameterized probability distributions, then $p$ can be chosen to make $G_i^{\cdot,\theta_0,\omega}(f, \beta_i)$ be the Fisher information matrix or energetic information matrix of $f(\cdot, \beta_i)$. In these cases, the set, $\mathcal{G}$, that the naturalized algorithms are covariant with respect to includes all parameterized probability distributions with positive definite Fisher information matrices and energetic information matrices.

Furthermore, Theorem 1 allows for the naturalization of a broad class of learning rules. For example, accelerated gradient methods [15] use updates of the form: $\beta_i = l_{i-1}(f, \theta_0, \omega) + \frac{i-1}{i+1}\left(l_{i-1}(f, \theta_0, \omega) - l_{i-2}(f, \theta_0, \omega)\right)$, $l_i = \beta_i - \alpha_{i-1} \int_{\mathcal{X}} \frac{\partial f}{\partial \beta_i}(\cdot, \beta_i) \, d\mu_i(f, \theta_0, \omega, \cdot)$, which can be transformed into a first-order covariant learning rule with respect to $(\beta_i)_{i=1}^\infty$ and $\mathcal{G}$ using Theorem 1. The resulting naturalized accelerated gradient update is: $l_i = \beta_i - \alpha_{i-1} \int_{\mathcal{X}} G_i^{\cdot,\theta_0,\omega}(f, \beta_i)^+ \frac{\partial f}{\partial \beta_i}(\cdot, \beta_i) \, d\mu_i(f, \theta_0, \omega, \cdot)$. Similarly, temporal difference algorithms [5] can be transformed into first order covariant learning rules. Dabney and Thomas [13] presented a naturalized temporal difference algorithm, which can be viewed as a successful application of Theorem 1, and which produced state of the art performances on several classical benchmark problems. Furthermore, since most discounted episodic policy gradient algorithms have been shown to *not* be gradient (or stochastic gradient) algorithms [19], natural policy gradient methods [12] are also examples of the application of Theorem 1 to non-gradient learning rules.

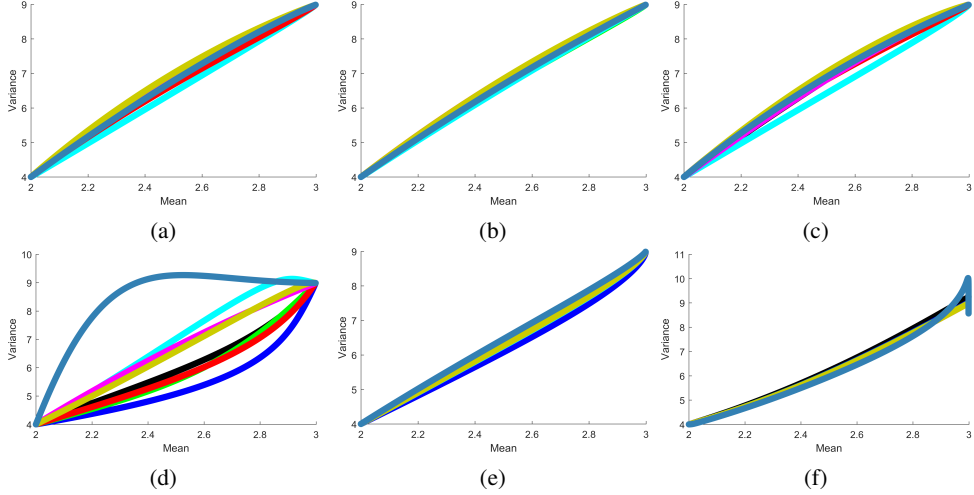

Figure 2: Reproduction of Figure 1 using naturalized gradient descent algorithms and with the legend suppressed. Each plot uses a fixed step size for all $k$, but step sizes vary between plots. **(a)** Where $f(x, \theta)$ is the log-probability of $x$ and $p_i(f, \theta_0, \omega, x, x) = f(x, \theta_i)$ so that $G_i^{z,\theta_0,\omega}(f, \beta_i)$ is the Fisher information matrix of $f(,\cdot, \beta_i)$. The Fisher information matrix was estimated from 1,000 samples of $x$. **(b)** The same as (a), except where $f(x,\theta)$ is the probability of $x$. **(c)** the same as (b), but using only 100 samples to estimate $G_i^{z,\theta_0,\omega}(f, \beta_i)$. **(e)** Same as (b), but using just 5 samples to estimate $G_i^{z,\theta_0,\omega}(f, \beta_i)$. **(e)** The same as (b), but where the $x$ used to estimate $G_i^{z,\theta_0,\omega}(f, \beta_i)$ were sampled from a continuous uniform distribution, $p_i$, rather than from $f(\cdot, \beta_i)$. **(f)** The same setup as (a), but using the direct estimation technique from Theorem 2.

Notice that the user of Theorem 1 is free to select what constitutes $f$ in an algorithm. For example, one might select $f$ to be the *probability density function* (PDF) for a normal distribution in the example from the introduction, or one might select $f$ to be the natural logarithm of the PDF for a normal distribution (this latter choice can make $G_i^{x,\theta_0,\omega}$ the Fisher information matrix).

Perhaps one of the most important properties of Theorem 1 is that $p_i(f, \theta_0, \omega, z, E, E')$ can have support only over some fixed small number of samples, $s > n$. This means that one can, for example, use a data-based estimate of the Fisher information matrix (constructed from $s$ samples), and the resulting update will be first-order covariant. The catch here is that $\mathcal{G}$ will be small for small $s$ (e.g., if $s < n$, then $\mathcal{G}$ will likely be empty). However, as $s$ grows, $\mathcal{G}$ will include more and more parameterizations of probability distributions until, in the limit as $s \to \infty$, $\mathcal{G}$ is the set of parameterized probability distributions whose Fisher information matrix is positive definite (the same $\mathcal{G}$ used implicitly in conventional covariance proofs [1]).

Lastly, notice that $p_i(f, \theta_0, \omega, z, E, E')$ can be unrelated to $\mu_i(f, \theta_0, \omega, E)$. This means, for example, that one can estimate the Fisher information matrix associated with one parameterized probability distribution using samples from a different distribution, and the resulting learning rule will be first-order covariant. This can be useful for applications where sampling from $f(\cdot, \theta)$ is expensive—for example in policy gradient applications where sampling states from the stationary distribution under a parameterized stochastic policy is expensive relative to sampling states from a uniform distribution. We empirically validated these various properties by applying various naturalized algorithms to the illustrative example from the introduction. The results, which support the theoretical discussion, are presented in Figure 2.

In summary, we have presented a method for converting a broad class of learning rules into first-order covariant learning rules. This method, which we refer to as the *naturalization* of a learning rule, extends work on natural gradient methods beyond gradient descent, and ensures covariance for metric tensors, $G_i^{x,\theta_0,\omega}$, that generalize the Fisher information matrix and energetic information matrix, without sacrificing covariance. We also showed how the updates produced by naturalized learning rules can be directly estimated, in some cases in linear time. Finally, we presented initial findings that suggest that there may not exist any practical second-order covariant learning rules.

# A  Proof of Theorem 1

We begin by establishing properties that we use later. Also, for brevity and to avoid clutter, we use several shorthand notations in all of the appendices: $G^x := G_i^{x,\theta_0,\omega}$, $\nabla \psi := \frac{\partial \psi}{\partial \beta_i}(\beta_i)$, $\nabla f := \frac{\partial f}{\partial \beta_i}(x, \beta_i)$, $\nabla^2 f := \frac{\partial^2 f}{\partial \beta_i^2}(x, \beta_i)$, $\nabla g := \frac{\partial g}{\partial \psi(\beta_i)}(x, \psi(\beta_i))$, and $\nabla^2 g := \frac{\partial^2 g}{\partial \psi(\beta_i)^2}(x, \psi(\beta_i))$.

**Property 1** (Jacobian Property). *If $f$ and $g$ are congruent representations, then for all $x \in \mathcal{X}$ and $\theta \in \mathbb{R}^n$,*

$$\frac{\partial f}{\partial \theta}(x, \theta) = \left(\frac{\partial \psi}{\partial \theta}(\theta)\right)^\intercal \frac{\partial g}{\partial \psi(\theta)}(x, \psi(\theta)).$$

*Proof.*

$$\frac{\partial f}{\partial \theta}(x, \theta) \stackrel{(a)}{=} \frac{\partial g}{\partial \theta}(x, \psi(\theta)) = \left(\frac{\partial \psi}{\partial \theta}(\theta)\right)^\intercal \frac{\partial g}{\partial \psi(\theta)}(x, \psi(\theta)),$$

where **(a)** holds because $f(x, \theta) = g(x, \psi(\theta))$ for all $x \in \mathcal{X}$ and $\theta \in \mathbb{R}^n$ by the assumption that $f$ and $g$ are congruent representations. □

**Property 2.** *For all parameterized functions, $f \in \mathcal{P}$, all $g \in \mathcal{P}$ that are congruent to $f$, all $x \in \mathcal{X}$, all $\theta_0 \in \Theta^\iota$, all $\omega \in \Omega$, and all $i \in \mathbb{N}_{>0}$,*

$$G^x(f, \beta_i) = \nabla \psi^\intercal G^x(g, \psi(\beta_i)) \nabla \psi.$$

*Proof.*

$$G^x(f, \beta_i) := \int_{\mathcal{X}^2} \frac{\partial f(x, \beta_i)}{\partial \beta_i} \frac{\partial f(y, \beta_i)}{\partial \beta_i}^\intercal dp_i(f, \theta_0, \omega, z, dx, dy)$$

$$\stackrel{(a)}{=} \int_{\mathcal{X}^2} \nabla \psi^\intercal \frac{\partial g(x, \psi(\beta_i))}{\partial \psi(\beta_i)} \frac{\partial g(y, \psi(\beta_i))}{\partial \psi(\beta_i)}^\intercal \nabla \psi \, dp_i(f, \theta_0, \omega, z, dx, dy)$$

$$\stackrel{(b)}{=} \nabla \psi^\intercal \int_{\mathcal{X}^2} \frac{\partial g(x, \psi(\beta_i))}{\partial \psi(\beta_i)} \frac{\partial g(y, \psi(\beta_i))}{\partial \psi(\beta_i)}^\intercal dp_i(g, \psi(\theta_0), \omega, z, dx, dy) \nabla \psi$$

$$= \nabla \psi^\intercal G^x(g, \psi(\beta_i)) \nabla \psi,$$

where **(a)** comes from Property 1 and **(b)** holds because $\nabla \psi$ does not depend on $x$ or $y$ and because $p$ is a covariant set of joint probability measures. □

**Property 3.** *If $l'$ is a first-order covariant update with respect to a sequence $(\beta_i)_{i=1}^\infty$, then for all $i \in \mathbb{N}_{>0}$, $\theta_0 \in \Theta^\iota$, and $\omega \in \Omega$,*

$$\nabla g^\intercal \nabla \psi \left(l'_i(f, \theta_0, \omega) - \beta_i\right) = \nabla g^\intercal \left(l'_i(g, \psi(\theta_0), \omega) - \psi(\beta_i)\right).$$

*Proof.* We begin by writing out the Taylor expansions in the definition of first-order covariance:

$$\tau_1(f(x, \cdot), \beta_i, l'_i(f, \theta_0, \omega)) = \tau_1(g(x, \cdot), \psi(\beta_i), l'_i(g, \psi(\theta_0), \omega))$$
$$f(x, \beta_i) + \nabla f^\intercal (l'_i(f, \theta_0, \omega) - \beta_i) = g(x, \psi(\beta_i)) + \nabla g^\intercal (l'_i(g, \psi(\theta_0), \omega) - \psi(\beta_i))$$
$$\nabla f^\intercal (l'_i(f, \theta_0, \omega) - \beta_i) \stackrel{(a)}{=} \nabla g^\intercal (l'_i(g, \psi(\theta_0), \omega) - \psi(\beta_i))$$
$$\nabla g^\intercal \nabla \psi (l'_i(f, \theta_0, \omega) - \beta_i) \stackrel{(b)}{=} \nabla g^\intercal (l'_i(g, \psi(\theta_0), \omega) - \psi(\beta_i)),$$

where **(a)** comes from the first terms on each side canceling by the definition of $\psi$ and **(b)** comes from Property 1. □

To establish Theorem 1 we show that for all $g \in \mathcal{G}$ that are congruent to $f$,

$$\tau_1(f(x, \cdot), \beta_i, \tilde{l}_i(f, \theta_0, \omega)) = \tau_1(g(x, \cdot), \psi(\beta_i), \tilde{l}_i(g, \psi(\theta_0), \omega)). \tag{3}$$

To establish (3), we write out the Taylor expansions, as in the proof of Property 3. This gives an equality which, if satisfied, implies that $\tilde{l}$ is first-order covariant with respect to $(\beta_i)_{i=1}^\infty$ and $\mathcal{G}$.

$$\tau_1(f(x,\cdot),\beta_i,\tilde{l}_i(f,\theta_0,\omega)) = \tau_1(g(x,\cdot),\psi(\beta_i),\tilde{l}_i(g,\psi(\theta_0),\omega))$$

$$f(x,\beta_i) + \nabla f^\intercal(\tilde{l}_i(f,\theta_0,\omega) - \beta_i) = g(x,\psi(\beta_i)) + \nabla g^\intercal(\tilde{l}_i(g,\psi(\theta_0),\omega) - \psi(\beta_i))$$

$$\nabla f^\intercal(\tilde{l}_i(f,\theta_0,\omega) - \beta_i) \stackrel{(a)}{=} \nabla g^\intercal(\tilde{l}_i(g,\psi(\theta_0),\omega) - \psi(\beta_i))$$

$$\nabla g^\intercal \nabla \psi (\tilde{l}_i(f,\theta_0,\omega) - \beta_i) \stackrel{(b)}{=} \nabla g^\intercal (\tilde{l}_i(g,\psi(\theta_0),\omega) - \psi(\beta_i)). \tag{4}$$

We will show that this condition is met.

$$\nabla g^\intercal \nabla \psi (\tilde{l}_i(f,\theta_0,\omega) - \beta_i)$$

$$\stackrel{(a)}{=} \nabla g^\intercal \nabla \psi \left( l'_i(f,\theta_0,\omega) - \beta_i + \int_\mathcal{X} G^\cdot(f,\beta_i)^+ \frac{\partial f}{\partial \beta_i}(\cdot,\beta_i)\, d\mu_i(f,\theta_0,\omega,\cdot) \right)$$

$$= \nabla g^\intercal \nabla \psi (l'_i(f,\theta_0,\omega) - \beta_i) + \nabla g^\intercal \nabla \psi \int_\mathcal{X} G^\cdot(f,\beta_i)^+ \frac{\partial f}{\partial \beta_i}(\cdot,\beta_i)\, d\mu_i(f,\theta_0,\omega,\cdot)$$

$$\stackrel{(b)}{=} \nabla g^\intercal (l'_i(g,\psi(\theta_0),\omega) - \psi(\beta_i)) + \nabla g^\intercal \nabla \psi \int_\mathcal{X} G^\cdot(f,\beta_i)^+ \frac{\partial f}{\partial \beta_i}(\cdot,\beta_i)\, d\mu_i(f,\theta_0,\omega,\cdot)$$

$$\stackrel{(c)}{=} \nabla g^\intercal (l'_i(g,\psi(\theta_0),\omega) - \psi(\beta_i)) + \nabla g^\intercal \nabla \psi \int_\mathcal{X} [\nabla \psi^\intercal G^\cdot(g,\psi(\beta_i)) \nabla \psi]^+ \frac{\partial f}{\partial \beta_i}(\cdot,\beta_i)\, d\mu_i(f,\theta_0,\omega,\cdot) j$$
(5)

where **(a)** comes from substituting in the definition of $\tilde{l}_i(f,\theta_{1:i},\omega)$, **(b)** holds by Property 3, and **(c)** comes from Property 2. Notice that $\nabla \psi^\intercal \in \mathbb{R}^{n\times m}$ has full column rank (since $m \le n$), $\nabla \psi \in \mathbb{R}^{m\times n}$ has full row rank, and $\text{rank}(G^x(g,\psi(\beta_i))) = m$ by the definition of $\mathcal{G}$ in Theorem 1. Thus, by Sylvester's rank inequality we have that $\text{rank}(G^x(g,\psi(\beta_i))\nabla \psi) \ge \text{rank}(G^x(g,\psi(\beta_i))) + \text{rank}(\nabla \psi) - m = m + m - m = m$. Also, due to its dimensions, $\text{rank}(G^x(g,\psi(\beta_i))\nabla \psi) \le m$, and so we can conclude that $\text{rank}(G^x(g,\psi(\beta_i))\nabla \psi) = m$. So, $\nabla \psi^\intercal$ has full column rank and $G^x(g,\psi(\beta_i))\nabla \psi$ has full row rank. So, by two applications of the rule that $(AB)^+ = B^+ A^+$ if $A$ has full column rank and $B$ has full row rank [20], we have that:

$$(\nabla \psi^\intercal G^\cdot(g,\psi(\beta_i))\nabla \psi)^+ = \nabla \psi^+ G^\cdot(g,\psi(\beta_i))^+ (\nabla \psi^\intercal)^+.$$

So, continuing (5), we therefore have that:

$$\nabla g^\intercal \nabla \psi (\tilde{l}_i(f,\theta_0,\omega) - \beta_i)$$

$$= \nabla g^\intercal (l'_i(g,\psi(\theta_0),\omega) - \psi(\beta_i)) + \nabla g^\intercal \nabla \psi \int_\mathcal{X} \nabla \psi^+ G^\cdot(g,\psi(\beta_i))^+ (\nabla \psi^\intercal)^+ \frac{\partial f}{\partial \beta_i}(\cdot,\beta_i)\, d\mu_i(f,\theta_0,\omega,\cdot)$$

$$= \nabla g^\intercal (l'_i(g,\psi(\theta_0),\omega) - \psi(\beta_i)) + \nabla g^\intercal \nabla \psi \nabla \psi^+ \int_\mathcal{X} G^\cdot(g,\psi(\beta_i))^+ (\nabla \psi^\intercal)^+ \frac{\partial f}{\partial \beta_i}(\cdot,\beta_i)\, d\mu_i(f,\theta_0,\omega,\cdot)$$

$$\stackrel{(a1)}{=} \nabla g^\intercal (l'_i(g,\psi(\theta_0),\omega) - \psi(\beta_i)) + \nabla g^\intercal \int_\mathcal{X} G^\cdot(g,\psi(\beta_i))^+ (\nabla \psi^\intercal)^+ \frac{\partial f}{\partial \beta_i}(\cdot,\beta_i)\, d\mu_i(f,\theta_0,\omega,\cdot)$$

$$\stackrel{(b)}{=} \nabla g^\intercal (l'_i(g,\psi(\theta_0),\omega) - \psi(\beta_i)) + \nabla g^\intercal \int_\mathcal{X} G^\cdot(g,\psi(\beta_i))^+ (\nabla \psi^\intercal)^+ \nabla \psi^\intercal \frac{\partial g}{\partial \psi(\beta_i)}(\cdot,\psi(\beta_i))\, d\mu_i(f,\theta_0,\omega,\cdot)$$

$$\stackrel{(a2)}{=} \nabla g^\intercal (l'_i(g,\psi(\theta_0),\omega) - \psi(\beta_i)) + \nabla g^\intercal \int_\mathcal{X} G^\cdot(g,\psi(\beta_i))^+ \frac{\partial g}{\partial \psi(\beta_i)}(\cdot,\psi(\beta_i))\, d\mu_i(f,\theta_0,\omega,\cdot)$$

$$\stackrel{(c)}{=} \nabla g^\intercal (l'_i(g,\psi(\theta_0),\omega) - \psi(\beta_i)) + \nabla g^\intercal \int_\mathcal{X} G^\cdot(g,\psi(\beta_i))^+ \frac{\partial g}{\partial \psi(\beta_i)}(\cdot,\psi(\beta_i))\, d\mu_i(g,\psi(\theta_0),\omega,\cdot)$$

$$= \nabla g^\intercal \left( \tilde{l}_i(g,\psi(\theta_0),\omega) - \psi(\beta_i) \right), \tag{6}$$

where **(a1)** and **(a2)** hold because $\nabla \psi$ has linearly independent rows because it is full rank, and has more columns than rows by the requirement that $m \le n$ in the definition of congruent representations, and so $\nabla \psi^+$ is a right-inverse, **(b)** holds by Property 1 and **(c)** holds be the assumption that $\mu$ is a covariant set of measures. Notice that (6) is equal to the right side of (4), and so we conclude.

# B   Derivation of Metric Tensor and Update Direction

In this appendix we provide intuition for what $\tilde l$ does relative to $l$, and show how $G_i^{x,\theta_0,\omega}$ encodes the dissimilarity function in (1). We begin by considering the stochastic gradient descent learning rule as an example to understand what the $\partial f(\cdot,\beta_i)/\partial\beta_i$ terms in a learning rule do, and how they should be changed to decouple the decisions of which learning rule to use and which parameterized function to use. For simplicity, in this appendix we consider the setting where $\beta_i := \theta_{i-1}$ and we use the shorthand $\theta_i := l_i(f,\theta_0,\omega)$. Also, recall that in all appendices we use the shorthands: $d := d_i^{x,\theta_0,\omega}$, and $G^x := G_i^{x,\theta_0,\omega}$.

The stochastic gradient descent update to make $f$ approximate some target function, $f^\star$, can be written as:
$$\theta_i = \theta_{i-1} + \alpha_i \underbrace{\left(f^\star(X_{i-1}(\omega)) - f(X_{i-1}(\omega),\theta_{i-1})\right)}_{=:\delta_{i-1}} \frac{\partial f}{\partial \theta_{i-1}}(X_{i-1}(\omega),\theta_{i-1}), \tag{7}$$

where $X_{i-1}:\Omega\to\mathcal{X}$ is a random variable, and where $(\alpha_i)_{i=1}^\infty$ is a sequence of small positive real-valued step sizes. For brevity, hereafter we write $X_{i-1}$ as shorthand for $X_{i-1}(\omega)$. In (7) the $\delta_{i-1}$ term is an *error term*. If $\delta_{i-1}$ is positive, it means that $\theta_i$ should be selected to make $f(X_{i-1},\theta_i)$ larger than $f(X_{i-1},\theta_{i-1})$. Similarly, if $\delta_{i-1}$ is negative, then it means that $\theta_i$ should be selected to make $f(X_{i-1},\theta_i)$ smaller than $f(X_{i-1},\theta_{i-1})$. This intuition is accomplished in (7) by multiplying $\delta_{i-1}$ by $\frac{\partial f}{\partial \theta_{i-1}}(X_{i-1},\theta_{i-1})$, which is a direction of change to $\theta_{i-1}$ that increases the value of $f(X_{i-1},\theta_{i-1})$.

However, there are many directions, $\Delta_{i-1}$, of change to the parameters, $\theta_{i-1}$, that would cause $f(X_{i-1},\theta_{i-1})$ to increase. In general, we could change the learning rule to be:
$$\theta_i = \theta_{i-1} + \delta_{i-1}\Delta_{i-1},$$
for any $\Delta_{i-1}$ such that (for infinitesimal $\alpha_i$) $f(X_{i-1},\theta_{i-1}+\alpha_i\Delta_{i-1}) \geq f(X_{i-1},\theta_{i-1})$. However, some directions, $\Delta_{i-1}$, are "better" than others. The error term, $\delta_{i-1}$, describes whether $f(X_{i-1},\theta_{i-1})$ should be bigger or smaller, but does not describe whether $f(x,\theta_{i-1})$ should be bigger or smaller for any $x \neq X_{i-1}$. Some directions, $\Delta_{i-1}$, might cause $f(X_{i-1},\theta_{i-1}+\alpha_i\Delta_{i-1})$ to increase slowly as $\alpha_i$ increases, but $f(x,\theta_{i-1}+\alpha_i\Delta_{i-1})$ to increase or decrease quickly as $\alpha_i$ increases, for some $x \neq X_{i-1}$. These $\Delta_{i-1}$ are not desirable because $\delta_{i-1}$ does not describe whether $f(x,\theta_{i-1})$ should be bigger or smaller. We desire a direction, $\Delta_{i-1}$, that does the opposite: it should cause $f(X_{i-1},\theta_{i-1}+\alpha_i\Delta_{i-1})$ to increase quickly with $\alpha_i$, and $f(x,\theta_{i-1}+\alpha_i\Delta_{i-1})$ to change slowly with $\alpha_i$ for all $x \neq X_{i-1}$.

We will focus our attention of the first constraint: we will find a direction, $\Delta_{i-1}$, that causes $f(X_{i-1},\theta_{i-1}+\alpha_i\Delta_{i-1})$ to increase *as quickly as possible* with $\alpha_i$. That is, we will select $\Delta_{i-1}$ to be a direction (vector of length one) such that for a step of infinitesimal length, $\alpha_i$, $f(X_{i-1},\theta_i+\alpha_i\Delta_{i-1})$ is maximized. More formally, we will select

$$\begin{aligned}
\Delta_{i-1} &:= \lim_{\alpha_i\to 0} \arg\max_{\Delta_{i-1}\in\{\Delta\in\mathbb{R}^n:\|\Delta\|=1\}} f(X_{i-1},\theta_{i-1}+\alpha_i\Delta_{i-1}) \\
&\overset{(a)}{=} \lim_{\alpha_i\to 0} \arg\max_{\Delta_{i-1}\in\{\Delta\in\mathbb{R}^n:\|\Delta\|=1\}} \frac{\partial f}{\partial \theta_{i-1}}(X_{i-1},\theta_{i-1})^\intercal(\alpha_i\Delta_{i-1}) + O(\alpha_i^2) \\
&= \lim_{\alpha_i\to 0} \arg\max_{\Delta_{i-1}\in\{\Delta\in\mathbb{R}^n:\|\Delta\|=1\}} \frac{\partial f}{\partial \theta_{i-1}}(X_{i-1},\theta_{i-1})^\intercal(\alpha_i\Delta_{i-1}) \\
&= \arg\max_{\Delta_{i-1}\in\{\Delta\in\mathbb{R}^n:\|\Delta\|=1\}} \frac{\partial f}{\partial \theta_{i-1}}(X_{i-1},\theta_{i-1})^\intercal\Delta_{i-1}, \tag{8}
\end{aligned}$$

where **(a)** comes from a Taylor expansion. By the method of Lagrange multipliers and the observation that $\|\Delta\|=1$ implies that $\|\Delta\|^2=1$, we have that any $\Delta_{i-1}$ that satisfies (8) must also satisfy:

$$\begin{aligned}
0 =& \frac{\partial}{\partial \Delta_{i-1}}\left(\frac{\partial f}{\partial \theta_{i-1}}(X_{i-1},\theta_{i-1})^\intercal\Delta_{i-1} - \lambda\left(\|\Delta_{i-1}\|^2 - 1\right)\right) \\
=& \frac{\partial}{\partial \Delta_{i-1}}\left(\frac{\partial f}{\partial \theta_{i-1}}(X_{i-1},\theta_{i-1})^\intercal\Delta_{i-1} - \lambda\left(\Delta_{i-1}^\intercal\Delta_{i-1} - 1\right)\right) \tag{9}\\
=& \frac{\partial f}{\partial \theta_{i-1}}(X_{i-1},\theta_{i-1}) - 2\lambda\Delta_{i-1},
\end{aligned}$$

and so
$$\Delta_{i-1} = \frac{1}{2\lambda} \frac{\partial f}{\partial \theta_{i-1}}(X_{i-1}, \theta_{i-1}),$$

where $\lambda$ is a scalar. It is straightforward to verify that this direction is the unique solution, $\Delta_{i-1}$, and not just a critical point of the Lagrangian. If we ignore the scalar terms (e.g., by viewing them as part of the step size, $\alpha_i$), we have the direction:

$$\Delta_{i-1} = \frac{\partial f}{\partial \theta_{i-1}}(X_{i-1}, \theta_{i-1}), \tag{10}$$

which is the direction of change to $\theta_{i-1}$ used by stochastic gradient descent in (7). It is also the direction of change to $\theta_{i-1}$ that is used by non-gradient learning rules, like temporal-difference learning [5], that use updates of the form:

$$\theta_i = \theta_{i-1} + \delta_{i-1}\Delta_{i-1},$$

where here $\delta_{i-1}$ denotes an error term called the *temporal difference error*. In general, for learning rules that satisfy Assumption 1, the $\partial f(\cdot, \theta_{i-1}i)/\partial \theta_{i-1}$ terms denote different $\Delta_{i-1}$ terms, evaluated using different $x \in \mathcal{X}$.

The problem with learning rules that use (10)—learning rules that satisfy Assumption 1—is that they make an implicit assumption that the distance between $f(X_{i-1}, \theta_{i-1})$ and $f(X_{i-1}, \theta_{i-1} + \alpha\Delta\theta_i)$ should be measured using Euclidean distance in the parameters when selecting $\Delta_{i-1}$. That is, they use $\|\Delta\|^2 \coloneqq \Delta^\intercal \Delta$ during the derivation of $\Delta_{i-1}$—specifically to obtain (9) during the derivation.

The problem with using learning rules that satisfy Assumption 1, which use Euclidean distance in the parameters when deriving $\Delta_{i-1}$, is that they intertwine the choices of which learning rule to use and which parameterized function to use. To see how this intertwining occurs, consider a parameterized function, $g$, that is congruent to $f$, with submersion $\psi$. Using $f$ and Euclidean distance in the parameterization, the squared distance between $f(X_{i-1}, \theta_{i-1})$ and $f(X_{i-1}, \theta_{i-1} + \Delta))$ is $\Delta^\intercal \Delta$. However, using $g$ and the Euclidean distance in the parameterization, the squared distance between the same two functions, $g(X_{i-1}, \psi(\theta_{i-1}))$ and $g(X_{i-1}, \psi(\theta_{i-1} + \Delta))$, is

$$(\psi(\theta_{i-1} + \Delta) - \psi(\theta_{i-1}))^\intercal (\psi(\theta_{i-1} + \Delta) - \psi(\theta_{i-1})),$$

which is not necessarily the same. These differing notions of distance will result in different solutions to (8), and thus different update directions. This is reflected by the fact that learning rules that satisfy Assumption 1 are not covariant or $j$-order covariant for any $j \in \mathbb{N}_{>0}$ and non-degenerate $\mathcal{G}$.

Furthermore, for some parameterizations, Euclidean distance in the parameters may be a poor notion of distance. For example, in a deep neural network, a weight at an early layer of the network may have little impact on the output of the network, while a weight near the output of the network might have a large impact. Using Euclidean distance in the parameters means that small changes to these two weights incur the same amount of distance, and so the direction of steepest ascent will favor larger changes to the weight later in the network, since small changes thereto can have a bigger influence on the network's output. Amari [6] was the first to suggest that this line of reasoning could explain the tendency of algorithms for training neural networks to require many iterations of the learning rule to properly set the values of weights early in the network.

This raises the question: what notion of distance (or more generally, what dissimilarity function) should be used when computing $\Delta_{i-1}$—the direction of steepest ascent of $f(X_{i-1}, \cdot)$ at $\theta_{i-1}$? We would like to use (1), so that

$$\|\Delta_{i-1}\|^2 \coloneqq \text{dist}(\theta_{i-1}, \theta_{i-1} + \Delta_{i-1})^2$$
$$= \frac{1}{2} \int_{\mathcal{X}^2} (f(x, \theta_{i-1}) - f(x, \theta_{i-1} + \Delta_{i-1}))(f(y, \theta_{i-1}) - f(y, \theta_{i-1} + \Delta_{i-1})p(dx, dy),$$

where $p(dx, dy) \coloneqq p_i(f, \theta_0, \omega, z, dx, dy)$, where $z \in \mathcal{X}$ corresponds to $z$ in (2). Although this definition of $\|\cdot\|$ is desirable, it does not ensure that a simple closed form exists for $\Delta_{i-1}$. So, instead we use

$$\|\Delta_{i-1}\|^2 \coloneqq \tau_2 \left(\text{dist}(\theta_{i-1}, \theta_{i-1} + \cdot)^2, \theta_{i-1}, \theta_{i-1} + \Delta_{i-1}\right).$$

That is, we use a second order Taylor approximation of the dissimilarity function, $d$ as our definition of squared distance. Although this second order Taylor approximation does *not* result in a definition of squared distance that yields covariant updates, Theorem 1 shows that it is sufficient to yield first-order covariant updates. Also, notice that this use a second order Taylor approximation to a dissimilarity function is not unprecedented: Amari's natural gradient method using the Fisher information matrix equates to using a second order Taylor approximation of Kullback–Leibler divergence to measure squared distances when computing $\Delta_{i-1}$ [14, Appendix A].

The use of a second-order Taylor approximation of $\text{dist}(\theta_{i-1}, \theta_{i-1} + \cdot)^2$ results in a closed form for the $\Delta_{i-1}$ that satisfy (8) because:

$$\tau_2\left(\text{dist}(\theta, \cdot)^2, \theta, \theta + \Delta\right) = \underbrace{\text{dist}(\theta, \theta)^2}_{=(a)} + \underbrace{\left(\frac{\partial \text{dist}}{\partial \gamma}(\alpha, \gamma)^2\big|_{\substack{\alpha=\theta \\ \gamma=\theta}}\right)^\intercal}_{=(b)} \Delta + \frac{1}{2}\Delta^\intercal \left(\frac{\partial^2 \text{dist}}{\partial \gamma^2}(\alpha, \gamma)^2\big|_{\substack{\alpha=\theta \\ \gamma=\theta}}\right) \Delta$$

$$= \frac{1}{2}\Delta^\intercal \left(\frac{\partial^2 \text{dist}}{\partial \gamma^2}(f, \alpha, \gamma)^2\big|_{\substack{\alpha=\theta \\ \gamma=\theta}}\right) \Delta,$$

since it is straightforward to verify that **(a)** and **(b)** are both zero.[1] Furthermore,

$$\left(\frac{\partial^2 \text{dist}}{\partial \gamma^2}(f, \alpha, \gamma)^2\big|_{\substack{\alpha=\theta \\ \gamma=\theta}}\right) = \int_{\mathcal{X}^2} \frac{\partial f}{\partial \theta}(x, \theta)\frac{\partial f}{\partial \theta}(y, \theta)^\intercal p(dx, dy).$$

So,

$$\tau_2\left(\text{dist}(\theta, \cdot)^2, \theta, \theta + \Delta\right) = \Delta^\intercal \left(\int_{\mathcal{X}^2} \frac{\partial f}{\partial \theta}(x, \theta)\frac{\partial f}{\partial \theta}(y, \theta)^\intercal p(dx, dy)\right) \Delta,$$

and thus

$$\|\Delta_{i-1}\|^2 := \Delta_{i-1}^\intercal G^x(f, \theta_{i-1}) \Delta_{i-1}.$$

Using this squared norm and the method of Lagrange multipliers as before, the solutions to (8) satisfy

$$0 = \frac{\partial}{\partial \Delta_{i-1}}\left(\frac{\partial f}{\partial \theta_{i-1}}(x, \theta_{i-1})^\intercal \Delta_{i-1} - \frac{1}{2}\lambda\left(\Delta_{i-1}^\intercal G^x(f, \theta_{i-1}) \Delta_{i-1} - 1\right)\right)$$

$$= \frac{\partial f}{\partial \theta_{i-1}}(x, \theta_{i-1}) - \lambda G^x(f, \theta_{i-1}) \Delta_{i-1},$$

and so $\Delta_{i-1} = \frac{1}{\lambda} G^x(f, \theta_{i-1})^+ \frac{\partial f}{\partial \theta_{i-1}}(x, \theta_{i-1})$, or ignoring the scalar terms as before (by viewing them as part of the step sizes),

$$\Delta_{i-1} = G^x(f, \theta_{i-1})^+ \frac{\partial f}{\partial \theta_{i-1}}(x, \theta_{i-1}).$$

This definition of $\Delta_{i-1}$ is exactly what is used by $\tilde{l}$.

## C  Proof of Theorem 2

Since $w^\star$ is a critical point:

$$0 = \int_{\mathcal{X}} (1 - \hat{1}(\cdot, w^\star))\frac{\partial \hat{1}}{\partial w^\star}(\cdot, w^\star)\, d\mu_i(f, \theta_0, \omega, \cdot)$$

$$0 = \int_{\mathcal{X}} \left(1 - (w^\star)^\intercal \frac{\partial f}{\partial \beta_i}(\cdot, \beta_i)\right)\frac{\partial f}{\partial \beta_i}(\cdot, \beta_i)\, d\mu_i(f, \theta_0, \omega, \cdot).$$

Rearranging terms, we obtain a new expression that is equal to a term in the learning rule, $l$:

$$\int_{\mathcal{X}} \frac{\partial f}{\partial \beta_i}(\cdot, \beta_i)\, d\mu_i(f, \theta_0, \omega, \cdot) = \int_{\mathcal{X}} \left((w^\star)^\intercal \frac{\partial f}{\partial \beta_i}(\cdot, \beta_i)\right)\frac{\partial f}{\partial \beta_i}(\cdot, \beta_i)\, d\mu_i(f, \theta_0, \omega, \cdot) \quad (11)$$

$$= \int_{\mathcal{X}} \frac{\partial f}{\partial \beta_i}(\cdot, \beta_i)\frac{\partial f}{\partial \beta_i}(\cdot, \beta_i)^\intercal\, d\mu_i(f, \theta_0, \omega, \cdot) w^\star, \quad (12)$$

---

[1]Notice that here we have switched notation for differentiation. This is because $\frac{\partial d}{\partial \theta_{i-1}}(\theta_{i-1}, \theta_{i-1})$ is ambiguous since the derivative is with respect to the second argument of $d$, not the first.

Replacing the left side of (11) in a learning rule, $l$, that satisfies Assumption 1, with (12), we have that $l$ can be written as:

$$l_i(f, \theta_0, \omega) = l'_i(f, \theta_0, \omega) + \left[\int_{\mathcal{X}} \frac{\partial f}{\partial \beta_i}(\cdot, \beta_i) \frac{\partial f}{\partial \beta_i}(\cdot, \beta_i)^{\mathsf{T}} \, d\mu_i(f, \theta_0, \omega, \cdot) \right] w^\star.$$

Similarly, $\tilde{l}$ from Theorem 1 can be written as

$$\tilde{l}_i(f, \theta_0, \omega) = l'_i(f, \theta_0, \omega) + \left[\int_{\mathcal{X}} G_i^{\cdot, \theta_0, \omega}(f, \beta_i)^+ \frac{\partial f}{\partial \beta_i}(\cdot, \beta_i) \frac{\partial f}{\partial \beta_i}(\cdot, \beta_i)^{\mathsf{T}} \, d\mu_i(f, \theta_0, \omega, \cdot) \right] w^\star. \quad (13)$$

Since

$$G_i^{x, \theta_0, \omega}(f, \beta_i) = \int_{\mathcal{X}} \frac{\partial f}{\partial \beta_i}(\cdot, \beta_i) \frac{\partial f}{\partial \beta_i}(\cdot, \beta_i)^{\mathsf{T}} \, d\mu_i(f, \theta_0, \omega, \cdot),$$

or

$$G_i^{x, \theta_0, \omega}(f, \beta_i) = \frac{\partial f}{\partial \beta_i}(x, \beta_i) \frac{\partial f}{\partial \beta_i}(x, \beta_i)^{\mathsf{T}},$$

and $G_i^{x, \theta_0, \omega}(f, \beta_i)$ is full rank, terms in (13) cancel to give:

$$\tilde{l}_i(f, \theta_0, \omega) = l'_i(f, \theta_0, \omega) + w^\star.$$

## D Proof of Theorem 3

We show that every learning rule, $l$, that is second-order covariant with respect to any sequence, $(\beta_i)_{i=1}^\infty$, and a set $\mathcal{G}$, must use the trivial update, $l_i(f, \theta_0, \omega) := \beta_i$ for all parameterized functions, $f$, where **1)** $n = k = 1$, **2)** both $g(x, \theta) := f(x, \ln(\theta))$ and $h(x, \theta) := f(x, \ln(\theta)/2)$ are in $\mathcal{G}$ and congruent to $f$ and **3)** both $\frac{\partial g}{\partial \theta}(\cdot, \beta_i)$ and $\frac{\partial^2 g}{\partial \theta^2}(\cdot, \beta_i)$ are not collinear and $\frac{\partial h}{\partial \theta}(\cdot, \beta_i)$ and $\frac{\partial^2 h}{\partial \theta^2}(\cdot, \beta_i)$ are not collinear.

To show this result, we will assume that $l$ is a second-order covariant learning rule and will then show that, under these conditions, $l_i(f, \theta_0, \omega) := \beta_i$. Since $l$ is second-order covariant with respect to $(\beta_i)_{i=1}^\infty$ and $\mathcal{G}$, we have that:

$$\tau_2(f(x, \cdot), \beta_i, l_i(f, \theta_0, \omega)) = \tau_2(g(x, \cdot), \psi(\beta_i), l_i(g, \psi(\theta_0), \omega)) = \tau_2(h(x, \cdot), \phi(\beta_i), l_i(h, \phi(\theta_0), \omega)),$$

and so:

$$a \nabla f + \frac{a^2}{2} \nabla^2 f = b \nabla g + \frac{b^2}{2} \nabla^2 g = c \nabla h + \frac{c^2}{2} \nabla^2 h, \quad (14)$$

where $a := l_i(f, \theta_0, \omega) - \beta_i$, $b := l_i(g, \psi(\theta_0), \omega) - \psi(\beta_i)$, $c := l_i(h, \phi(\theta_0), \omega) - \phi(\beta_i)$, $\nabla h := \frac{\partial h}{\partial \phi(\beta_i)}(x, \phi(\beta_i))$, and $\nabla^2 h := \frac{\partial^2 h}{\partial \phi(\beta_i)^2}(x, \phi(\beta_i))$.

We will show that, given $f$ and the $g$ and $h$ specified in the theorem, (14) is only satisfied by $a = 0$, $b = 0$, and $c = 0$, which by the definitions of $a$, $b$, and $c$ implies our result. Specifically, let:

$$g(\psi(\theta)) := f(\ln(\psi(\theta)))$$
$$h(\phi(\theta)) := f\left(\frac{1}{2}\ln(\phi(\theta))\right).$$

So, $g$ and $h$ are congruent to $f$ with submersions $\psi(\theta) = e^\theta$ and $\phi(\theta) = e^{2\theta}$, respectively. Thus, we have the following:

$$\nabla \psi = e^{\beta_i}$$
$$\nabla \phi = 2e^{2\beta_i}$$
$$\nabla^2 \psi = e^{\beta_i}$$
$$\nabla^2 \phi = 4e^{2\beta_i}$$
$$\nabla f = \nabla \psi \nabla g = e^{\beta_i} \nabla g$$
$$\nabla f = \nabla \phi \nabla h = 2e^{2\beta_i} \nabla h$$
$$\nabla^2 f = \nabla^2 g \nabla \psi^2 + \nabla g \nabla^2 \psi = e^{2\beta_i} \nabla^2 g + e^{\beta_i} \nabla g$$
$$\nabla^2 f = \nabla^2 h \nabla \phi^2 + \nabla h \nabla^2 \phi = 4e^{4\beta_i} \nabla^2 h + 4e^{2\beta_i} \nabla h.$$

From (14) we have the requirement that for all $x \in \mathcal{X}$:

$$\begin{aligned}
b\nabla g + \frac{b^2}{2}\nabla^2 g &= a\nabla f + \frac{a^2}{2}\nabla^2 f \\
&= ae^{\beta_i}\nabla g + \frac{a^2}{2}\left(e^{2\beta_i}\nabla^2 g + e^{\beta_i}\nabla g\right) \\
&= \left(ae^{\beta_i} + \frac{a^2}{2}e^{\beta_i}\right)\nabla g + \frac{a^2}{2}e^{2\beta_i}\nabla^2 g.
\end{aligned} \qquad (15)$$

Recall that $\nabla g$ and $\nabla^2 g$ (and $\nabla h$ and $\nabla^2 h$) are not collinear functions. Thus, the only way for (15) to hold for all $x$ is if

$$b = ae^{\beta_i} + \frac{a^2}{2}e^{\beta_i}, \qquad (16)$$

and

$$\frac{b^2}{2} = \frac{a^2}{2}e^{2\beta_i}. \qquad (17)$$

Similarly, from (14) we have the requirement that for all $x \in \mathcal{X}$:

$$\begin{aligned}
c\nabla h + \frac{c^2}{2}\nabla^2 h &= a\nabla f + \frac{a^2}{2}\nabla^2 f \\
&= 2ae^{2\beta_i}\nabla h + \frac{a^2}{2}\left(4e^{4\beta_i}\nabla^2 h + 4e^{2\beta_i}\nabla h\right) \\
&= \left(2ae^{2\beta_i} + \frac{a^2}{2}4e^{2\beta_i}\right)\nabla h + a^2 2e^{4\beta_i}\nabla^2 h,
\end{aligned}$$

and thus we have that

$$c = 2ae^{2\beta_i} + \frac{a^2}{2}4e^{2\beta_i}, \qquad (18)$$

and

$$\frac{c^2}{2} = a^2 2e^{4\beta_i}. \qquad (19)$$

It is straightforward to verify using a computer algebra system like Wolfram Alpha that the only values for $a$, $b$, $c$ that satisfy (16), (17), (18), and (19) simultaneously occur when $a = b = c = 0$. Since $a = b = c = 0$ corresponds to the trivial learning rule, $l_i(f, \theta_0, \omega) = \beta_i$, we conclude.